\documentclass[letterpaper]{article} 
\usepackage{aaai25}  
\usepackage{times}  
\usepackage{helvet}  
\usepackage{courier}  
\usepackage[hyphens]{url}  
\usepackage{graphicx} 
\usepackage{natbib}  
\usepackage{caption} 
\usepackage{algorithm}
\usepackage{algorithmic}
\usepackage{booktabs}
\usepackage{xlop}
\usepackage{amsmath}
\usepackage{multirow}

\usepackage{newfloat}
\usepackage{listings}
\DeclareCaptionStyle{ruled}{labelfont=normalfont,labelsep=colon,strut=off} 
\floatstyle{ruled}
\newfloat{listing}{tb}{lst}{}
\floatname{listing}{Listing}
\def\eg{\emph{e.g.}}

\title{Dissecting Multiplication in Transformers: Insights into LLMs}
\author{
    Luyu Qiu\textsuperscript{\rm 1},
    Jianing Li\textsuperscript{\rm 2},
    Chi Su\textsuperscript{\rm 2},
    Chen Jason Zhang\textsuperscript{\rm 2},
    Lei Chen\textsuperscript{\rm 1}
}
\affiliations{
    \textsuperscript{\rm 1}Hong Kong University of Science and Technology\\
    \textsuperscript{\rm 2}Hong Kong Polytechnic University\\

    lqiuag@connect.ust.hk, {tensor.li,chisu, jason-c.zhang}@polyu.edu.hk, leichen@cse.ust.hk
}

\usepackage{bibentry}

\begin{document}

\maketitle

\begin{abstract}
Transformer-based large language models have achieved remarkable performance across various natural language processing tasks. However, they often struggle with seemingly easy tasks like arithmetic despite their vast capabilities. This stark disparity raise human's concerns about their safe and ethical use, hinder their widespread adoption.
In this paper, we focus on a typical arithmetic task, integer multiplication, to explore and explain the imperfection of transformers in this domain. We provide comprehensive analysis of a vanilla transformer trained to perform $n$-digit integer multiplication. Our observations indicate that the model decomposes multiplication task into multiple parallel subtasks, sequentially optimizing each subtask for each digit to complete the final multiplication.
Based on observation and analysis, we infer the reasons of transformer’s deficiencies in multiplication tasks lies in their difficulty in calculating successive carryovers and caching intermediate results, and confirmed this inference through experiments. Guided by these findings, we propose improvements to enhance transformer's performance on multiplication tasks. These enhancements are validated through rigorous testing and mathematical modeling, not only enhance transformer's interpretability, but also improve its performance, \eg, we achieve over 99.9\% accuracy on 5-digit integer multiplication with a tiny transformer, outperform LLMs GPT-4. 
Our method contributes to the broader fields of model understanding and interpretability, paving the way for analyzing more complex tasks and Transformer models. This work underscores the importance of explainable AI, helping to build trust in large language models and promoting their adoption in critical applications.
\end{abstract}

%

\section{Introduction}
Recent years witnessed rapid development of Natural Language Process (NLP). Transformers~\cite{vaswani2017attention}, the core of modern Large Language Models (LLMs), have achieved remarkable performance across various tasks include dialogue, chatting, and translation~\cite{vaswani2017attention,devlin2018bert,raffel2020exploring,touvron2023llama,dziri2024faith}.
The rise and success of recent Large Language Models (LLM) like GPT-4~\cite{achiam2023gpt} and Gemini~\cite{team2023gemini} have profoundly impacted various domains in our society, highlighting new possibilities in Artificial General Intelligence (AGI).

\begin{figure}
\centering
\includegraphics[width=0.85\linewidth]{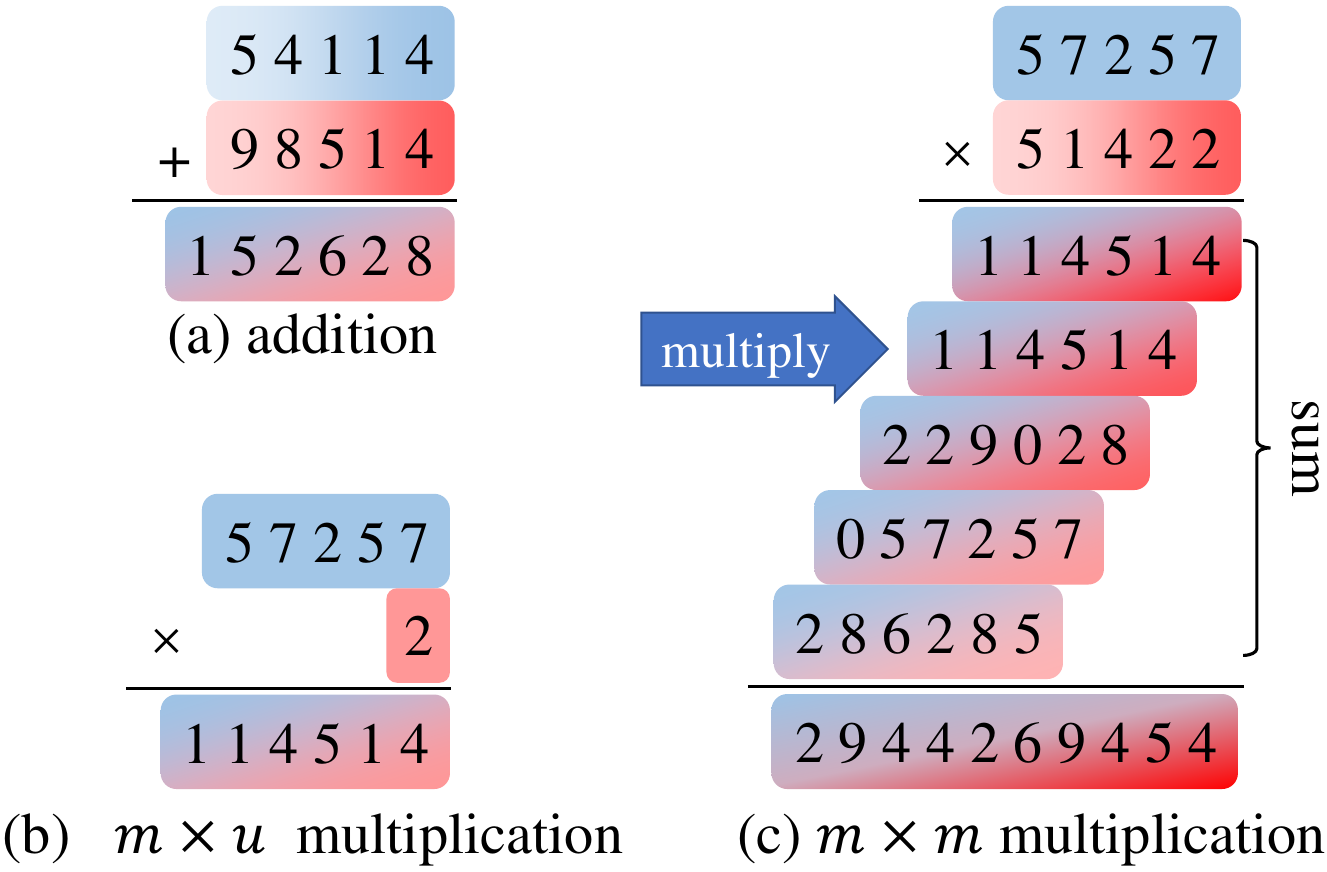}\\
\caption{The decomposed steps of (a) addition, (b) multi-digit and unit-digit ($m\times u$) multiplication, and (c) multi-digit and multi-digit ($m\times m$) multiplication.}
\label{fig:multi}
\vspace{-2mm}
\end{figure}


In stark contrast, when we attempt to apply transformers to various applications, we encounter difficulties with arithmetic problems, \eg, the modern LLM GPT-4~\cite{achiam2023gpt} even struggles with tasks like simple integer multiplication~\cite{dziri2024faith}, a basic calculation that is easy for human to perform.

Table~\ref{table:gpt4acc} shows the accuracy of GPT-4 on  integer multiplication tasks.
The stark disparity between the remarkable achievements of LLMs on seemingly complex tasks and their unexpected shortcomings on trivial ones raise human’s concerns about their safe and ethical use, and prompts critical questions about their capabilities. 

Seeking thorough answers to these questions remains an open research challenge. In this paper, we offer novel insights into the fundamental limits of transformers, centered around arithmetic problem.
Arithmetic tasks shows several differences from typical natural language tasks. The primary distinction lies in following aspects:
1) Data format: in typical NLP tasks, characters are sequentially arranged from front to back. Whereas in arithmetic problem like addition or multiplication, calculation starts from high digit may increases the difficulty.
2) Intermediate step: arithmetic calculation involves complex intermediate calculation steps, how to accurately process the intermediate result is important for transformer. 
3) The uniqueness: typical NLP tasks allow diverse outcomes and the model encourages diversified outputs, while arithmetic calculations usually have definite results, diversified outputs may hurt the prediction accuracy.
4) Data lack: the proportion of arithmetic data in internet-collected training dataset is low, and the data is used only 1 $\mathtt{\sim}$ 2 epoch for training, which may not sufficient for training. 
These characteristics hinder the efficiency and accuracy of transformers in arithmetic tasks. Nonetheless, the unique nature of arithmetic tasks provides an excellent opportunity to understand and reveal the internal mechanisms of transformer models. In this paper, we focus on the insufficiency of transformer in arithmetic problems and investigate possible explanations.

The advent of eXplainable AI (XAI) enable researchers and practitioners to understand the reason behind models' predictions, and has significantly mitigated the issues of opacity and lack of interpretability inherent in advanced machine learning models~\cite{shapley1951notes,petsiuk2018rise,chang2019explaining}.
By unraveling the internal mechanisms of these models, XAI helps us comprehend their strengths, limitations, and potential failure points, thereby facilitating the safe and ethical use, and the development of more reliable systems.
Transformers have demonstrated exceptional performance across a wide range of tasks. However, their inherent complexity, such as multi-layered self-attention architecture, makes it difficult to understand how they generate specific outputs~\cite{vig2019multiscale}.
Current transformer-related XAI methods primarily focus on  interpretability for specific samples in particular tasks, such as indirect object identification ~\cite{wang2022interpretability} and Colored Objects\cite{merullo2023circuit}. These methods provide valuable insights into the decision-making processes of these models. Nonetheless, as tasks grow more complex, the intrinsic correlations within these tasks and the overall decision-making processes are often overlooked. This oversight limits the effectiveness of existing XAI techniques when applied on transformers.

Previous works~\cite{quirke2023understanding} analyze a single-layer transformer model on integer addition task, providing a template for examining arithmetic tasks with transformers. While addition is a relatively simple task with few intermediate steps, which cannot fully leverage the capabilities of transformers. 
In this paper, we focus on dissecting more complex multiplication task. Fig.~\ref{fig:multi} visualizes the intermediate steps of addition and multiplication. From the figure, we can see that addition shares similar complexity with multiplication with unit-digit multiplier. While the complexity of multiplication task increases significantly when extending multiplier to multi-digit.
We hence initiated our analysis on a simplified version of multiplication, \eg, multi-digit $\times$ unit-digit (omitted as $m\times u$), and then extend analysis on universal multi-digit $\times$ multi-digit ($m\times m$) multiplication.

We show that the transformer separates the multiplication task into independent subtasks, which are computed in parallel. Different subtasks are employed to predict the digits in different position in answer numbers. Observations on loss curves indicate a sequential learning order for these subtasks. 
We inferred the model's deficiencies in handling multiplication tasks based on its attention patterns, and confirmed these through experiments. Our findings provide valuable insights into understanding and interpreting transformers. These insights may also have broader implications for AI safety and reliability.
The contribution of this work can be summarized as follows,
\begin{itemize}
\item We thoroughly dissect the multiplication task within transformers, and provide detailed explanation of transformer's incapacity on multiplication task, which are confirmed through experiments.
\item We proposed targeted enhancements to refine transformer's incapacity in multiplication. Equipped with these enhancements, we achieve 99.9\% accuracy with tiny transformer.
\item The proposed analysis pipeline contributes to the broader fields of model understanding and interpretability, paving the way for analyzing
more complex tasks and Transformer models.
\end{itemize}

\begin{table}
\caption{The accuracy (\%) of GPT-4 on 1-5 digits multiplications. We use the prompt “What's the answer of $a * b$?”. The row and column represent the digits of multiplicand and multiplier, respectively. Each combination of multiplicand and multiplier is tested 100 times.}
\setlength{\tabcolsep}{12 pt}
\small
\begin{tabular}{c|ccccc}
\toprule
\# Digits &1   &2   &3   &4   &5 \\\midrule
1         &100 &100 &100 &99  &100 \\\midrule
2         &100 &97  &95  &74  &59  \\\midrule
3         &100 &91  &60  &25  &17  \\\midrule
4         &99  &82  &29  &6   &1   \\\midrule
5         &94  &58  &13  &0   &0   \\\bottomrule
\end{tabular}
\label{table:gpt4acc}
\vspace{-2mm}
\end{table}

\section{Related Work}
This work is closely related to transformer and  eXplainable AI (XAI). This section briefly summarizes those two categories of works.

\subsection{Transformer }
Transformers~\cite{vaswani2017attention} model long-range dependencies among linguistic tokens through self-attention mechanism. Due to the outstanding performance in sequence modeling tasks, transformers have been applied to a wide range of tasks in natural language processing~\cite{devlin2018bert,brown2020language,taylor2022galactica,thoppilan2022lamda,chung2024scaling} and computer vision~\cite{radford2021learning,li2023blip,li2024mini}.
Recent studies try to use transformers to solve arithmetic problems. ~\cite{qian2022limitations} find the limitation of language models in arithmetic tasks, and propose to introduce fine-grained computation steps to mitigate the limitations of transformer. 
~\cite{yang2023gpt} concentrate on improving the reasoning process and teach the model step-by-step. Based on this ideology, they trained a model MathGLM, which consists of 2B parameters to perform arithmetic task and outperform GPT4 on several arithmetic problems, But the performance is still not perfect. 
~\cite{lee2023teaching} focused on the data format design in arithmetic problems. Their finding reveals that using detailed, instructive data with intermediate steps will improve accuracy and sample complexity.
~\cite{dziri2024faith} investigate the limitations of transformer models in representative compositional tasks such as multi-digit multiplication. They formulate compositional tasks as computation graphs to systematically quantify the level of complexity and find that transformer's performance rapidly decay with increased task complexity. 
Existing methods analyze transformer's shortcomings in arithmetic tasks based solely on accuracy, without deeply understanding the internal mechanisms of how transformers compute these tasks. Moreover, the proposed multi-step decomposition refinements significantly reduces the training and inference efficiency.

\subsection{XAI in Transformer}
The eXplainable AI (XAI) aims to interpret and reverse engineering neural networks and transformers to find meaningful circuits, and has been an area of active research.
Recently, numerous XAI studies have emerged to explain the decision-making processes of transformers across various tasks
~\cite{wang2022interpretability,merullo2023circuit,quirke2023understanding}.
~\cite{wang2022interpretability} identify the function of different attention head and group the attention heads into 7 main classes to to perform indirect-object identification task. 
~\cite{merullo2023circuit} propose to adjust the attention heads in middle layers to ‘repair’ the transformer's prediction in Colored Objects task.
~\cite{quirke2023understanding} provide a comprehensive analysis
of a one-layer transformer on n-digit integer addition task through decomposing addition into three base functions and analyse the learning condition of each function.
~\cite{shen2023positional} find the reliance on positional information cause the transformer model get poor performance on arithmetic problems with a small number of digits, and propose corresponding refinement.
Existing XAI methods for arithmetic tasks have proposed various improvements, significantly enhancing accuracy. However, they lack in-depth analysis of how models conduct these tasks. This paper provides a detailed analysis of the model's computation process, visually demonstrating how transformers handle arithmetic tasks and proposing improvements to address identified incapabilities.

\section{Method}
\subsection{Background}
In this paper, we focus on the integer multiplication task within transformer. The input to transformer is a sequence of words or symbols, which consists of two $n$-digit numbers $D=(D_{n-1},...,D_1,D_0)$ and $D'=(D_{n-1}',...,D_1',D_0')$, along with operators. Table~\ref{table:data_format} shows formats of input data. The model first converts the input into a sequence of one-hot vectors representing the corresponding symbols through a vocabulary table of size $V$. 
These one-hot vectors are then mapped to a sequence of embeddings, $\textbf{x}=(x_1, x_2, ..., x_L)$, where $x_i\in \mathcal{R}^d$ is the embedding of $i$-th word with dimension $d$, $L$ is the sequence length. 

The transformer processes the input embeddings using a series of blocks that consist of Multi-head Self-Attention (MSA) and Feed-Forward (FF) layers. Initially, the input embeddings are combined with positional embeddings:
\begin{equation}
\begin{aligned}
x_i = x_i+\mathbf{E}^{pos}_i.
\end{aligned}
\end{equation}
The combined embeddings are then passed through MSA layer, which captures the importance of each embedding relative to the others and producing contextually enriched embeddings.
These enriched embedding are subsequently fed into FF layer,
\begin{equation}
\begin{aligned}
\textbf{x} &= \operatorname{MSA}(\textbf{x})+\textbf{x},\\
\textbf{x} &= \operatorname{FF}(\textbf{x})+\textbf{x}.
\end{aligned}
\end{equation}
Finally, the output embeddings are mapped back to words through vocabulary table.

\begin{table}
\caption{Examples of the data format for multiplication. The symbol after ``$=$" is what the model needs to predict.}
\setlength{\tabcolsep}{10 pt}
\small
\begin{tabular}{c|c}
\toprule
Data Format       &Example  \\\midrule
$m\times u$ mul   &$57257\times 2=114514$      \\\midrule
Reversed $m\times u$ mul   &$57257\times 2=415411$      \\\midrule
$m\times m$ mul   &$57257\times 51422=2944269454$      \\\midrule
Reversed $m\times m$ mul   &$57257\times 51422=4549624492$      \\\midrule
\end{tabular}
\label{table:data_format}
\end{table}

\subsection{Multiplication}
The integer multiplication task has a vast problem and solution space. For 5-digit integer multiplication, this space exceeds 10 billion ($100,000 \times 100,000$). Moreover, unlike addition task~\cite{quirke2023understanding}, multiplication involves a series of intermediate steps, especially when dealing with large numbers. These complex intermediate steps hinder transformers from  predicting accurate products. Even the leading language model GPT-4~\cite{achiam2023gpt}, finds it challenging to handle extensive multiplications (See Table~\ref{table:gpt4acc}).
In Fig.\ref{fig:multi}, we visualize the intermediate steps of humans completing addition and multiplication tasks. 
From the figure, we observe that the multi-digit $\times$ multi-digit ($m \times m$) multiplication is decomposed into several multi-digit $\times$ unit-digit ($m \times u$) multiplications followed by subsequent addition tasks. We begin investigation with $m \times u$ multiplication to simplify analysis, and then extend to $m \times m$ multiplication.


\begin{figure*}
\centering
\includegraphics[width=1\linewidth]{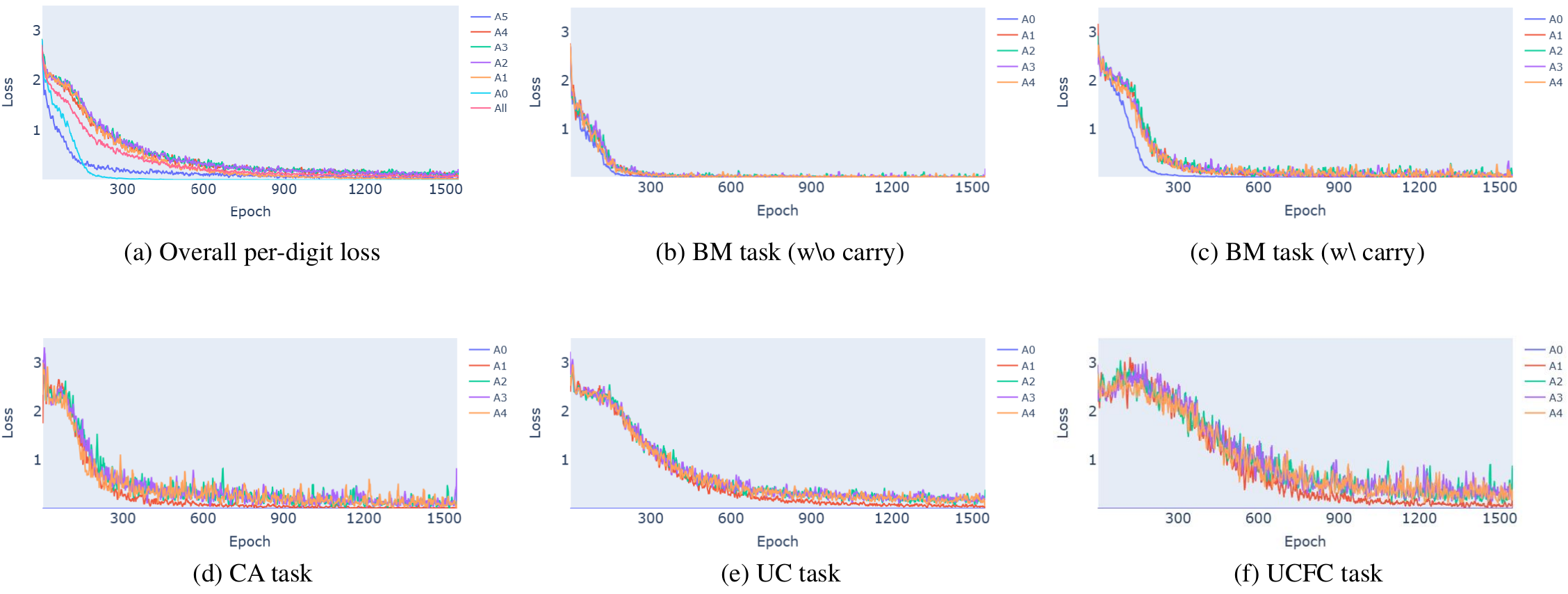}\\
\caption{Illustrations of (a) the overall per-digit loss curve, and (b-f) per-digit loss curve for each subtask.}
\label{fig:overall_loss}
\end{figure*}

\subsection{Unit-digit Multiplication Analysis}
In this section, we analyze transformers on $m \times u$ multiplication task. Mirroring the way humans calculate multiplication, we assert that the transformer utilizes several basic subtasks to complete multiplication.

\begin{itemize}
\item \textbf{Base Multiply (BM)}: BM calculates the product of two single digits $d_i$ and $d_j'$ at each position. Based on whether produce carry to next position, BM can be divided into BM (w\textbackslash carry) and BM (w\textbackslash o carry).
\item \textbf{Carry (CA)}: CA is responsible for calculating the carryover from the previous digit.
\item \textbf{Use Carry (UC)}: UC considers the carry from previous digit and add the carry to the product at current position.
\end{itemize}
Based on above basic subtasks, transformer can chain multiple subtasks together to achieve complex functionalities, \eg, \textbf{Use Carry and Further Carry (UCFC)} considers carry from previous column, and further produces carry to next column.

\subsubsection{Overall Analysis:}
We first investigate how transformer learns each subtask.
Fig.~\ref{fig:overall_loss}(a) shows per-digit training loss for each subtask and overall task on 5-digit  $m\times u$ multiplication. In the figure, ``A$n$" denotes the $n$-th digit in answer numbers, ``$\operatorname{All}$" denote the overall loss curve for all digits. 

The loss curves in Fig.~\ref{fig:overall_loss}(a) show transformer learns each answer digit semi-independently. 
Digit A0 (units digit) and digit A5 are learnt much more quickly than other digits and with less noise.
This is expected because the calculation of A0, the units digit, does not need to consider carry, making it simpler than other digits.
In contrast to a0, the calculation of A5 only needs to consider the carry.
The loss curves of A1-A4 exhibit similar decline patterns, because the calculation complexity of these digit is comparable, \eg, they all need to add the carry from previous column to the product at current column, and further make carry to next column.

\subsubsection{Per-task Analysis:}

We then analyse each subtask in Fig.~\ref{fig:overall_loss}(b-f). We categorise the training data into non-overlapping subsets aligned to each subtask and illustrate per-digit loss curves for each subtask. To better analysis, we visualize the training curves of BM (w\textbackslash carry) and BM (w\textbackslash o carry) separately. 
Overviewing Fig.~\ref{fig:overall_loss}(b-f), we can find an interesting pattern. At the beginning, all tasks have high loss. As the training goes on, all loss curves drop but there is a ``time lag" between the loss drop of different tasks. 
The curve of BM(w\textbackslash o carry) drops faster than all other curves. This is expected because BM(w\textbackslash o carry) is the basic of all other tasks, it must be accurate before subsequent calculations. 
The second fastest declining losses are BM (w\textbackslash carry) and Carry loss. These two subtask are learned simultaneously because they are complementary calculation. The convergence of these two task indicates that the model is able to accurately calculate carry.
The UC loss exhibits a significant plateau phase at the beginning of the training. 
This plateau phase matches our expectation that UC subtask requires accurate calculation of Carry, hence UC loss can only decrease after the Carry task converges.
UCFC loss decreases the latest, because the computation of UCFC spans 3 digits, \eg, product at current digit, carry from previous digit, and make carry to next digit. Therefore, its convergence occurs after all other losses converge.

From the figure, we can also observe that, UCFC's loss curve  is much noisier than others. This is because the calculation of UCFC spans 3 digits, inaccuracy in any single digit will lead to errors in UCFC. 
We find that transformer has relatively low loss on simple UCFC cases (\eg, the multiplier is $0$ or $1$). This is because when the multiplier is $0$ or $1$, the answer is 0 or the multiplicand itself, reduce calculation difficulty. While transformer encounters high loss value when multiplier is other digits, which usually lead to cascaded UCFC case, \eg, $47134\times 9=424206$.

\subsubsection{Reserve Answer Digit:}
Usually humans calculate multiplication from lower digits. Inspired by this, recent studies try to enhance transformer's accuracy in arithmetic through reversing answer digits~\cite{shen2023positional,lee2023teaching}. The data formats of reversed digits are shown in Table~\ref{table:data_format}. In this section, we validated the effectiveness of reversing answer digits and analyzed the principles behind its success.

We first compare the accuracy of transformer trained on ordinal and reversed formats. We denote the transformer trained with ordinal and reversed data formats as ``ordinal transformer" and ``reversed transformer, respectively. , The results are summarise in Table ~\ref{table:head}
In the table, the reversed transformer consistently achieves superior accuracy across different numbers of attention heads.
This is because reversed transformer starting calculations from lower digits, hence can take advantage of previously generated answer digits.
When the number of attention head in transformer greater exceeds 3, the performance improvement tends to saturate. This indicates that 3 attention heads are sufficient to complete multiplication.

\begin{table}
\caption{Accuracy (\%) of ordinal and reversed format with different attention heads.}
\setlength{\tabcolsep}{25 pt}
\small
\begin{tabular}{c|cc}
\toprule
\#Heads   &Ordinal &Reversed  \\\midrule
1        &66.2  &83.6  \\\midrule
2        &85.3  &96.3  \\\midrule
3        &89.8  &99.3  \\\midrule
4        &90.1  &99.3  \\\midrule
5        &91.1  &100  \\\midrule
6        &91.2  &99.9  \\\midrule
\end{tabular}
\label{table:head}
\vspace{-2mm}
\end{table}

\begin{figure}
\centering
\includegraphics[width=1\linewidth]{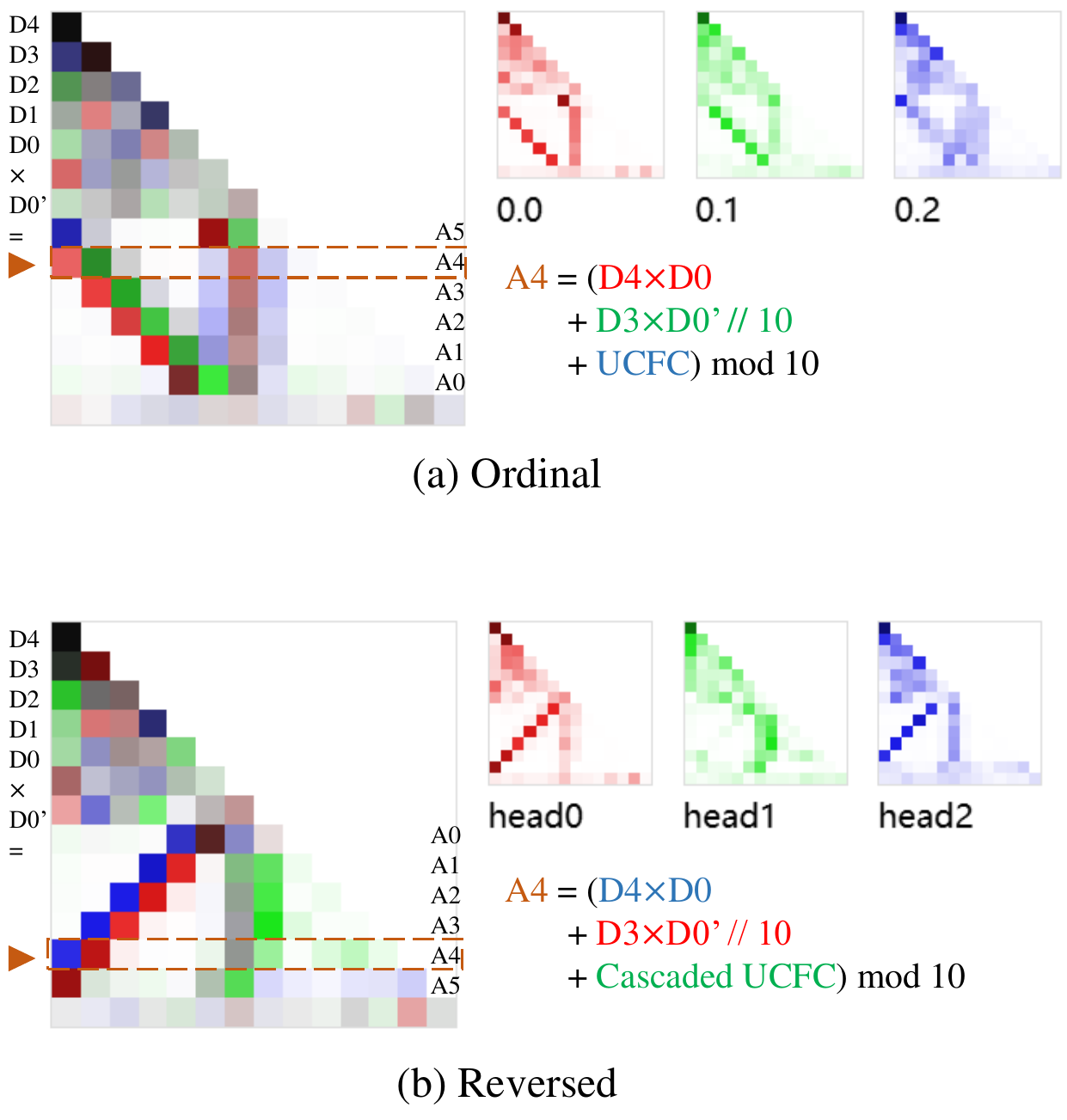}\\
\caption{Attention map of ordinal and reversed answer digit format. To predict answer digit, the multiple attention heads are responsible for different tasks and combine the information in subsequent MLP layers.}
\label{fig:attn}
\end{figure}

Next, we investigate the reasons behind the superiority of reversed transformer. We visualize the attention heads of transformer with heads in Fig.~\ref{fig:attn}. 
In Fig.~\ref{fig:attn}, ``D4" to ``=" in the left column denote the input tokens, and ``A0" to ``A5" in the right part of attention map denote the predicted answer.
Before ``=" token is revealed, the model has not seen full question and every digit token is independent, the attention maps don't show regular pattern.
The model starts to calculate multiplication at $8$-th row (corresponding to ``=" token), and the calculate is completed at $13$-row, the last row is not used. 
Both heads in ordinal and reversed transformers show a apparent “staircase” pattern, but with different directions. This is expected because the staircase pattern means the heads attend to each pair of digits to be multiplied. 

We first analyze ordinal transformer. In Fig.~\ref{fig:attn}(a), the heads attend to digits in multiplicand sequentially from left to right, and 3 attention heads have 1-token offset from each other.
We find that each attention head is responsible for different subtasks. For example, to calculate digits A4, the red attention head calculate \textbf{BM} on digit D4, the green head focus on \textbf{Carry} at D3, The blue head be responsible for…the \textbf{UCFC} computation at D3 (cross D2 to D4 digits). The information from 3 heads is combined in the subsequent MLP layer to generate the final answer digit A4.
While the calculation of first answer digit A5 does not need \textbf{BM} of red head, and last digit A0 does not consider \textbf{Carry} from previous column.
The attention map also point out the reason why transformer struggles with handling cascaded UCFC task: transformer only focus 3 continuous tokens in input sequence, while the calculation of cascaded UCFC need at least 4 tokens.
We then validated the function of each head through ablation experiments. When we invalidate the red head, the loss of BM task increase significantly. While ablate green and blue head has little influence on BM task. Similar conclusions can be observed in the other subtasks as well. This indicates that each attention head has a well-defined role.

\begin{table*}
\caption{Overall and per-digit accuracy (\%) of ordinal and reversed transformer with different multiplier formats. }
\setlength{\tabcolsep}{2.2 pt}
\small
\begin{tabular}{c|c|cccccccccc|c|ccccccccccc}
\toprule
Multiplier &\multicolumn{11}{|c}{Ordinal}  &\multicolumn{11}{|c}{Reversed} \\\cmidrule{2-23}
Format &overall     &A9&A8&A7&A6&A5&A4&A3&A2&A1&A0 &overall    &A9&A8&A7&A6&A5&A4&A3&A2&A1&A0 \\\midrule
$0000d$  &85.3&100 &100 &100 &100 &98.4&95.4&94.4&94.9&99.6&100    &99.3&100 &100 &100 &100 &99.9&99.8&99.8&99.8&99.9&100 \\\midrule
$d000d$ &16.8&98.6&96.9&95.2&85.1&40.2&28.6&96.0&96.4&99.5&100    &22.0&100 &99.8&99.8&96.2&42.5&34.4&99.9&99.9&100 &100 \\\midrule
$000dd$ &5.2 &100 &100 &100 &93.8&45.2&21.6&19.6&20.0&32.0&100    &8.1 &100 &100 &100 &99.9&40.6&20.8&20.4&21.7&40.5&100\\\midrule
$00ddd$ &0.2 &100 &100 &94.3&43.1&13.4&11.3&10.8&12.4&31.6&100    &0.4 &100 &100 &99.9&36.8&11.5&10.9&10.6&12.9&40.0&100\\\midrule 
$0dddd$ &0   &100 &95.2&44.4&11.8&10.0&10.4&9.3 &12.4&31.4&100    &0.1 &100 &98.7&32.5&11.2& 9.9&10.5&10.2&13.6&38.0&100\\\midrule 
$ddddd$ &0   &91.3&39.9&13.4&10.5&9.9 &9.5 &10.5&12.3&32.6&100    &0   &99.5&32.5&11.9&10.5&10.4&9.5 &10.5&13.4&36.3&100\\\midrule 
\end{tabular}
\label{table:digitacc}
\end{table*}

We then focus on reversed transformer. In Fig.~\ref{fig:attn}(b) the reversed transformer exhibits opposite attention pattern, the model attends to digits in a staircase pattern from right to left. 
Each attention head in reversed transformer is also responsible for different subtasks. The blue head is responsible for \textbf{BM} calculation. The red head calculate the \textbf{carry} from previous column. Different from ordinal transformer, the reversed transformer could access to previously generated answer digits, hence green head mainly focus on previous answer digits and is responsible for calculating cascaded carry.
This attention patterns also explains why the reversed transformer performs better: the reversed transformer can better utilize the already generated digits to calculate the remaining results, hence can better handle the cascaded UCFC cases.

Finally, we investigate transformers with fewer or more than 3 attention heads. We found that in the transformer with less than 3 heads, each head handles plural subtasks. Conversely, when there are more than 3 heads, multiple heads end up performing the same function.

\subsection{Multi-digit Multiplication Analysis}
In this section, we extend analysis to $m\times m$ multiplication task. When we conducting $m\times m$ multiplication, we encountered a significant accuracy drop with either ordinal or reversed transformer. 
In following part, we analyzed the reasons for the transformer's incapacity in the $m\times m$ multiplication task and propose potential solutions.

\subsubsection{Analysis:}
We first investigate per-digit accuracy of ordinal and reversed transformers on $m\times m$ multiplication task with different multiplier formats. The experimental results are summarised in  Table~\ref{table:digitacc}. 
The first column in Table~\ref{table:digitacc} denotes the multiplier format, \eg, ``$0000d$" denotes a 5-digit number with only unit-digit unfixed, \eg, $d\in\{0,1,...,9\}$,  while other digits fixed as zero value. And ``$000dd$" denotes the unit-digit and tens-digit are not fixed. All the multiplicands are regular five-digit numbers.

\begin{figure}
\centering
\includegraphics[width=1\linewidth]{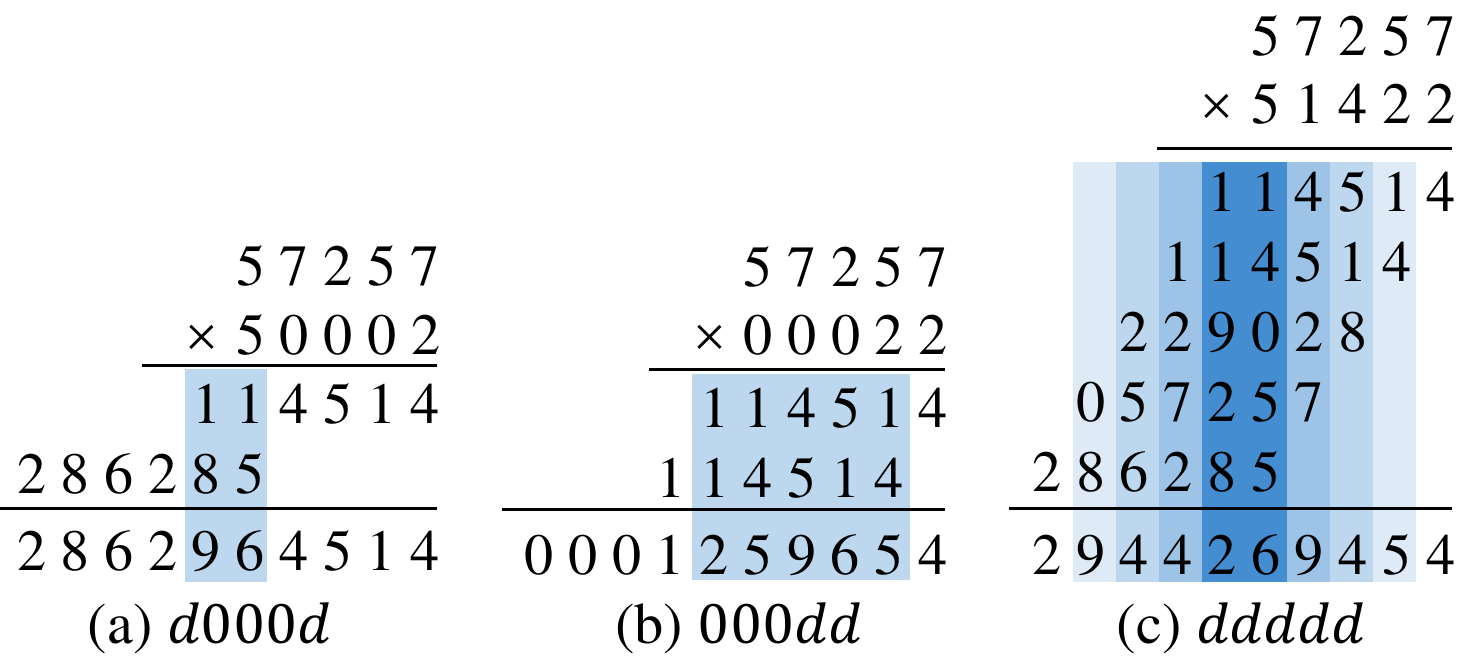}\\
\caption{The overlap of per-digit product with different multiplier format. The darker the color, the more overlapping digits there are.}
\label{fig:overlap}
\end{figure}

It can be observed in Table~\ref{table:digitacc} that, using multiplier with only unit digit non-fixed, \eg, $0000d$, both ordinal and reversed transformers achieve high accuracy at all answer digits A0-A9. 
The intermediate digits exhibit slightly lower accuracy, \eg, A1-A5, this is because these intermediate digits require extra carry calculation, which is not exist in unit digit A0. The table also shows that reversed transformer achieves better accuracy on digit A1-A5, indicating its superiority in carry calculation.

We then explored the cases where the multiplier has more than one non-fixed digits, which corresponding 2-6 row in Table~\ref{table:digitacc}. 
In the table, we observed that when there is an overlap between the intermediate products of multiplicand and each digit in multiplier, the accuracy of answer digits decreases. And the more digits overlap at the digit, the lower the accuracy at that digit. For example, in $ddddd\times ddddd$, the digits A4 and A5 need to consider all intermediate products, hence have the lowest accuracy. 
Fig.~\ref{fig:overlap} visualize the overlap of intermediate products with different multiplier formats, the darker color denotes more overlap exists in this position.
In Fig.~\ref{fig:overlap}(a), A4 and A5 in answer digits have overlap from 2 per-digit products, which corresponding to the accuracy decrease of A4 and A5 of $2$-nd row in Table~\ref{table:digitacc}.  The darker color in Fig.~\ref{fig:overlap}(c) also match the accuracy decrease in the last row of Table~\ref{table:digitacc}.
This indicates that single-layer transformer lacks the capacity to handle the multiple intermediate step in $m\times m$ multiplication task.

Additionally, when we apply the transformer trained on $m\times m$ task to $m\times u$ task, we suffer severe accuracy drop. This indicates directly train transformer on challenging $m\times m$ task prevents it from learning even basic multiplication rules, hence resulting in poor performance on simpler $m\times u$ multiplication task.

\subsubsection{Refinement:}

Based on above analysis, we identified the reasons for the inferior accuracy of transformer on $m\times m$ multiplication task:
1) The ordinal answer digit format increase the difficulty for the transformer to calculate carries.
2) The $m\times m$ task involves multiple steps of calculation, requiring deeper model to store intermediate results. 
3) Directly using complex $m\times m$ data for training increases the learning difficulty for the model.

We addressed above limitations to enhance transformer's ability on arithmetic task in following aspects, 
1) Reversing the answer digits to enable the transformer can better calculate the carry.
2) Increase model depth to handle more intermediate steps.
3) Increase the proportion of simple samples during training to allow the model to learn progressively. (The simple sample is defined as multiplier with only 1 digit unfix, while other digits are fixed as 0).

\begin{table}
\caption{Effectiveness of each refinement on $m\times m$ multiplication task.}
\setlength{\tabcolsep}{20 pt}
\small
\begin{tabular}{l|c}
\toprule
Method           &Acc.  \\\midrule
baseline           &0.0   \\\midrule
baseline + reverse &0.0   \\\midrule
baseline + depth   &1.6   \\\midrule
baseline + sample  &0.0   \\\midrule
baseline + depth + reverse &8.7   \\\midrule
baseline + depth + sample  &64.9   \\\midrule
baseline + depth + reverse + sample  &99.9   \\\midrule
\end{tabular}
\label{table:refine}
\vspace{-2mm}
\end{table}

\begin{table}
\caption{Accuracy (\%) of ordinal and reversed transformer with different depth.}
\setlength{\tabcolsep}{8 pt}
\small
\begin{tabular}{c|cc|cc}
\toprule
\multirow{2}{*}{\# Layers} &\multicolumn{2}{|c|}{$m\times u$}  &\multicolumn{2}{|c}{$m\times m$} \\
\cmidrule{2-5}
 &Ordinal &Reversed &Ordinal &Reversed  \\\midrule
1         &85.3 &99.3  &0.0  &0.0   \\\midrule
2         &92.3 &99.9  &0.0  &33.6  \\\midrule
4         &92.7 &99.8  &7.5  &97.4  \\\midrule
8         &95.5 &99.8  &64.9 &99.6  \\\midrule
12        &95.1 &99.9  &79.1 &99.9  \\\midrule
\end{tabular}
\label{table:reversed_acc}
\vspace{-2mm}
\end{table}

We first verify the effectiveness of above 3 refinement and the experimental results are summarised in Table~\ref{table:refine}. In the table, ``reverse" refers to reversing the answer digits, ``depth" denotes increase the transformer depth, and ``sample" denotes increase the proportion of simple samples.
In the table, solely ``reverse" or ``sample" also bring no promotion. This is because the model's capacity is not enough for the complex intermediate steps. Increasing model depth bring limited accuracy promotion. 
When we combine ``depth" and ``reverse", we observed a slightly higher accuracy, \eg, 8.7\%. This shows deep model trained with reverse format is beneficial. Increase the proportion of simple samples bring significant accuracy promotion. This shows  progressive learning reduce the overall training difficulty. When we combine all 3 refinements, we achieve remarkable accuracy. This experiment indicates the effectiveness of 3 improvements.

\subsubsection{Depth Analysis:}
In this part, we further analyzed the impact of model depth. 
Table~\ref{table:reversed_acc} shows the accuracy of transformer models with different layers on $m\times u$ and $m\times m$ multiplication tasks, respectively.
The single-layer transformer only performs well on $m\times u$ multiplication task, but it suffers failure on more complex $m\times m$ task. This is because single-layer model's capacity is insufficient to handle intermediate steps.
Increasing model depth bring significant accuracy promotion on $m\times m$ task for both ordinal and reversed transformer. 
The performance promotion of reverse transformer began to saturate at depth of 4, while ordinal transformer does not reach saturation even at depth of 12. This is because the ordinal model cannot utilize the previously generated answer digits, hence the calculation of each digit needs to consider more information and involves more intermediate step, necessitating a deeper model.

\subsubsection{Simple Sample Proportion:}
\begin{figure}
\centering
\includegraphics[width=1\linewidth]{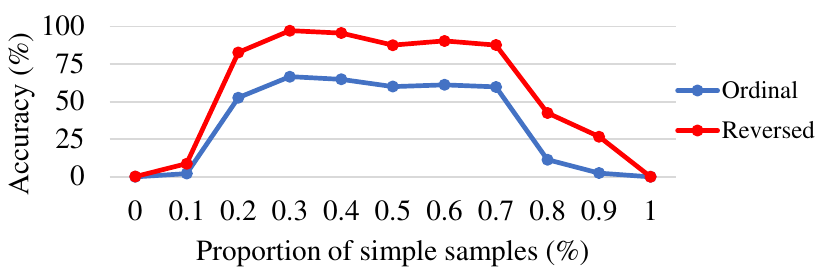}\\
\caption{Accuracy (\%) of ordinal and reversed transformer
trained with different proportion of simple samples.}
\label{fig:prop}
\vspace{-2mm}
\end{figure}

We then investigate the influence of simple sample proportion. We visualize the accuracy of ordinal and reversed transformers in Fig.~\ref{fig:prop}. The x-axis represents the proportion of simple samples. From the figure, we can see that both too many and too few simple samples are detrimental to model's learning. Too few simple samples make it difficult for the model to learn the basic rules of multiplication, while too many simple samples cause the model lack the ability to handle more difficult ones. This highlights the importance of setting an appropriate proportion of simple samples.

\subsection{Analysis of LLMs' incapability in Arithmetic}
Based on experiments and analysis in this paper, we tried to analyze to possible reasons for the suboptimal performance of current LLMs on arithmetic tasks. 

\begin{itemize}
\item Data format: existing LLMs are trained on  internet-collected data, in which the digits are usually presented in ordinal order, increases models' learning difficulty.
\item Data lack: the internet-collected training data contains few arithmetic samples, and most of the data is used for only 1-2 epochs. Moreover, the proportion of simple data is insufficient. This makes it difficult for LLMs to learn the basic rules of arithmetic tasks progressively from easy to difficult.
\item Output diversity : typical NLP tasks allow more diverse outcomes and the model encourages diversified outputs, while arithmetic calculations usually have definite results.
\item Tokenizer: in some LLMs' tokenizer processes, multiple characters are packaged as a single token , \eg, in GPT-4, the input arithmetic formula ``$12345\times 857$" is divided into ``123", ``45", ``$\times$" and ``857". This tokenizer further increases the difficulty for LLMs.

\end{itemize}

\subsection{Implementation Details}
\textbf{Training:} We start analysis with single layer decoder-only transformer on $m\times u$ task and extend to multi-layer transformer on $m\times m$ task. Our training consists of 2000 iterations with batch size 64. Each digit in training data is sampled independently from a uniform distribution \{0,1,...,9\}. We use Adam optimizer and the learning rate is set as 1e-4. 

\textbf{Inference:} During inference, the input consists of multiplicand, times sign, multiplier and equal sign, \eg, ``$12345\times 54321=$". The model generate each digit of product answer in an autoregressive manner. The autoregressive process stops when generating the final digit. We test on 1 million randomly sampled data which not appear in training set.

\section{Conclusion}
In this paper, we focused on the limitations of transformers in arithmetic problems and propose a integrated pipeline to elucidate the internal mechanism of transformer.
We visually analyzed the process of how transformers perform multiplication tasks. Our experiments show that transformers decompose the multiplication task into multiple parallel subtasks. The learning of these subtasks follows a certain order and each subtask is handled by different attention heads. Our 
We found that the reason transformers struggle with arithmetic tasks and propose targeted solutions to address these issues.
The proposed explainability pipeline contributes to the broader fields of model understanding and interpretability, we believe it will pave
the way for analyzing more complex tasks and transformer models.

\clearpage

\bibliography{aaai25}

\begin{thebibliography}{25}
\providecommand{\natexlab}[1]{#1}

\bibitem[{Achiam et~al.(2023)Achiam, Adler, Agarwal, Ahmad, Akkaya, Aleman,
  Almeida, Altenschmidt, Altman, Anadkat et~al.}]{achiam2023gpt}
Achiam, J.; Adler, S.; Agarwal, S.; Ahmad, L.; Akkaya, I.; Aleman, F.~L.;
  Almeida, D.; Altenschmidt, J.; Altman, S.; Anadkat, S.; et~al. 2023.
\newblock Gpt-4 technical report.
\newblock \emph{arXiv preprint arXiv:2303.08774}.

\bibitem[{Brown et~al.(2020)Brown, Mann, Ryder, Subbiah, Kaplan, Dhariwal,
  Neelakantan, Shyam, Sastry, Askell et~al.}]{brown2020language}
Brown, T.; Mann, B.; Ryder, N.; Subbiah, M.; Kaplan, J.~D.; Dhariwal, P.;
  Neelakantan, A.; Shyam, P.; Sastry, G.; Askell, A.; et~al. 2020.
\newblock Language models are few-shot learners.
\newblock \emph{Advances in neural information processing systems}, 33:
  1877--1901.

\bibitem[{Chang et~al.(2019)Chang, Creager, Goldenberg, and
  Duvenaud}]{chang2019explaining}
Chang, C.-H.; Creager, E.; Goldenberg, A.; and Duvenaud, D. 2019.
\newblock Explaining Image Classifiers by Counterfactual Generation.
\newblock arXiv:1807.08024.

\bibitem[{Chung et~al.(2024)Chung, Hou, Longpre, Zoph, Tay, Fedus, Li, Wang,
  Dehghani, Brahma et~al.}]{chung2024scaling}
Chung, H.~W.; Hou, L.; Longpre, S.; Zoph, B.; Tay, Y.; Fedus, W.; Li, Y.; Wang,
  X.; Dehghani, M.; Brahma, S.; et~al. 2024.
\newblock Scaling instruction-finetuned language models.
\newblock \emph{Journal of Machine Learning Research}, 25(70): 1--53.

\bibitem[{Devlin et~al.(2018)Devlin, Chang, Lee, and
  Toutanova}]{devlin2018bert}
Devlin, J.; Chang, M.-W.; Lee, K.; and Toutanova, K. 2018.
\newblock Bert: Pre-training of deep bidirectional transformers for language
  understanding.
\newblock \emph{arXiv preprint arXiv:1810.04805}.

\bibitem[{Dziri et~al.(2024)Dziri, Lu, Sclar, Li, Jiang, Lin, Welleck, West,
  Bhagavatula, Le~Bras et~al.}]{dziri2024faith}
Dziri, N.; Lu, X.; Sclar, M.; Li, X.~L.; Jiang, L.; Lin, B.~Y.; Welleck, S.;
  West, P.; Bhagavatula, C.; Le~Bras, R.; et~al. 2024.
\newblock Faith and fate: Limits of transformers on compositionality.
\newblock \emph{Advances in Neural Information Processing Systems}, 36.

\bibitem[{Lee et~al.(2023)Lee, Sreenivasan, Lee, Lee, and
  Papailiopoulos}]{lee2023teaching}
Lee, N.; Sreenivasan, K.; Lee, J.~D.; Lee, K.; and Papailiopoulos, D. 2023.
\newblock Teaching arithmetic to small transformers.
\newblock \emph{arXiv preprint arXiv:2307.03381}.

\bibitem[{Li et~al.(2023)Li, Li, Savarese, and Hoi}]{li2023blip}
Li, J.; Li, D.; Savarese, S.; and Hoi, S. 2023.
\newblock Blip-2: Bootstrapping language-image pre-training with frozen image
  encoders and large language models.
\newblock In \emph{International conference on machine learning}, 19730--19742.
  PMLR.

\bibitem[{Li et~al.(2024)Li, Zhang, Wang, Zhong, Chen, Chu, Liu, and
  Jia}]{li2024mini}
Li, Y.; Zhang, Y.; Wang, C.; Zhong, Z.; Chen, Y.; Chu, R.; Liu, S.; and Jia, J.
  2024.
\newblock Mini-gemini: Mining the potential of multi-modality vision language
  models.
\newblock \emph{arXiv preprint arXiv:2403.18814}.

\bibitem[{Merullo, Eickhoff, and Pavlick(2023)}]{merullo2023circuit}
Merullo, J.; Eickhoff, C.; and Pavlick, E. 2023.
\newblock Circuit component reuse across tasks in transformer language models.
\newblock \emph{arXiv preprint arXiv:2310.08744}.

\bibitem[{Petsiuk, Das, and Saenko(2018)}]{petsiuk2018rise}
Petsiuk, V.; Das, A.; and Saenko, K. 2018.
\newblock Rise: Randomized input sampling for explanation of black-box models.
\newblock \emph{arXiv preprint arXiv:1806.07421}.

\bibitem[{Qian et~al.(2022)Qian, Wang, Li, Li, and Yan}]{qian2022limitations}
Qian, J.; Wang, H.; Li, Z.; Li, S.; and Yan, X. 2022.
\newblock Limitations of language models in arithmetic and symbolic induction.
\newblock \emph{arXiv preprint arXiv:2208.05051}.

\bibitem[{Quirke et~al.(2023)}]{quirke2023understanding}
Quirke, P.; et~al. 2023.
\newblock Understanding addition in transformers.
\newblock \emph{arXiv preprint arXiv:2310.13121}.

\bibitem[{Radford et~al.(2021)Radford, Kim, Hallacy, Ramesh, Goh, Agarwal,
  Sastry, Askell, Mishkin, Clark et~al.}]{radford2021learning}
Radford, A.; Kim, J.~W.; Hallacy, C.; Ramesh, A.; Goh, G.; Agarwal, S.; Sastry,
  G.; Askell, A.; Mishkin, P.; Clark, J.; et~al. 2021.
\newblock Learning transferable visual models from natural language
  supervision.
\newblock In \emph{International conference on machine learning}, 8748--8763.
  PMLR.

\bibitem[{Raffel et~al.(2020)Raffel, Shazeer, Roberts, Lee, Narang, Matena,
  Zhou, Li, and Liu}]{raffel2020exploring}
Raffel, C.; Shazeer, N.; Roberts, A.; Lee, K.; Narang, S.; Matena, M.; Zhou,
  Y.; Li, W.; and Liu, P.~J. 2020.
\newblock Exploring the limits of transfer learning with a unified text-to-text
  transformer.
\newblock \emph{Journal of machine learning research}, 21(140): 1--67.

\bibitem[{Shapley(1951)}]{shapley1951notes}
Shapley, L.~S. 1951.
\newblock Notes on the n-person game—ii: The value of an n-person game.

\bibitem[{Shen et~al.(2023)Shen, Bubeck, Eldan, Lee, Li, and
  Zhang}]{shen2023positional}
Shen, R.; Bubeck, S.; Eldan, R.; Lee, Y.~T.; Li, Y.; and Zhang, Y. 2023.
\newblock Positional description matters for transformers arithmetic.
\newblock \emph{arXiv preprint arXiv:2311.14737}.

\bibitem[{Taylor et~al.(2022)Taylor, Kardas, Cucurull, Scialom, Hartshorn,
  Saravia, Poulton, Kerkez, and Stojnic}]{taylor2022galactica}
Taylor, R.; Kardas, M.; Cucurull, G.; Scialom, T.; Hartshorn, A.; Saravia, E.;
  Poulton, A.; Kerkez, V.; and Stojnic, R. 2022.
\newblock Galactica: A large language model for science.
\newblock \emph{arXiv preprint arXiv:2211.09085}.

\bibitem[{Team et~al.(2023)Team, Anil, Borgeaud, Wu, Alayrac, Yu, Soricut,
  Schalkwyk, Dai, Hauth et~al.}]{team2023gemini}
Team, G.; Anil, R.; Borgeaud, S.; Wu, Y.; Alayrac, J.-B.; Yu, J.; Soricut, R.;
  Schalkwyk, J.; Dai, A.~M.; Hauth, A.; et~al. 2023.
\newblock Gemini: a family of highly capable multimodal models.
\newblock \emph{arXiv preprint arXiv:2312.11805}.

\bibitem[{Thoppilan et~al.(2022)Thoppilan, De~Freitas, Hall, Shazeer,
  Kulshreshtha, Cheng, Jin, Bos, Baker, Du et~al.}]{thoppilan2022lamda}
Thoppilan, R.; De~Freitas, D.; Hall, J.; Shazeer, N.; Kulshreshtha, A.; Cheng,
  H.-T.; Jin, A.; Bos, T.; Baker, L.; Du, Y.; et~al. 2022.
\newblock Lamda: Language models for dialog applications.
\newblock \emph{arXiv preprint arXiv:2201.08239}.

\bibitem[{Touvron et~al.(2023)Touvron, Lavril, Izacard, Martinet, Lachaux,
  Lacroix, Rozi{\`e}re, Goyal, Hambro, Azhar et~al.}]{touvron2023llama}
Touvron, H.; Lavril, T.; Izacard, G.; Martinet, X.; Lachaux, M.-A.; Lacroix,
  T.; Rozi{\`e}re, B.; Goyal, N.; Hambro, E.; Azhar, F.; et~al. 2023.
\newblock Llama: Open and efficient foundation language models.
\newblock \emph{arXiv preprint arXiv:2302.13971}.

\bibitem[{Vaswani et~al.(2017)Vaswani, Shazeer, Parmar, Uszkoreit, Jones,
  Gomez, Kaiser, and Polosukhin}]{vaswani2017attention}
Vaswani, A.; Shazeer, N.; Parmar, N.; Uszkoreit, J.; Jones, L.; Gomez, A.~N.;
  Kaiser, {\L}.; and Polosukhin, I. 2017.
\newblock Attention is all you need.
\newblock \emph{Advances in neural information processing systems}, 30.

\bibitem[{Vig(2019)}]{vig2019multiscale}
Vig, J. 2019.
\newblock A multiscale visualization of attention in the transformer model.
\newblock \emph{arXiv preprint arXiv:1906.05714}.

\bibitem[{Wang et~al.(2022)Wang, Variengien, Conmy, Shlegeris, and
  Steinhardt}]{wang2022interpretability}
Wang, K.; Variengien, A.; Conmy, A.; Shlegeris, B.; and Steinhardt, J. 2022.
\newblock Interpretability in the wild: a circuit for indirect object
  identification in gpt-2 small.
\newblock \emph{arXiv preprint arXiv:2211.00593}.

\bibitem[{Yang et~al.(2023)Yang, Ding, Lv, Jiang, He, Guo, Bai, and
  Tang}]{yang2023gpt}
Yang, Z.; Ding, M.; Lv, Q.; Jiang, Z.; He, Z.; Guo, Y.; Bai, J.; and Tang, J.
  2023.
\newblock Gpt can solve mathematical problems without a calculator.
\newblock \emph{arXiv preprint arXiv:2309.03241}.

\end{thebibliography}

\end{document}